\documentclass[journal]{IEEEtran}
\usepackage{booktabs}
\usepackage{multirow}
\usepackage{algorithm}
\usepackage[noend]{algpseudocode}
\usepackage{amsmath}
\usepackage{tabularx}
\usepackage{amsthm}

\usepackage{graphicx}
\usepackage{amsthm}
\usepackage{authblk}
\usepackage{makecell}
\graphicspath{ {images/} }
\usepackage{fancyhdr}
\usepackage[dvipsnames]{xcolor}
\usepackage{tikz}  
\usetikzlibrary{arrows,shapes,chains}  
\ifCLASSINFOpdf
\else
\fi
\begin{document}

\title{Smart Power Supply for UAV Agility Enhancement Using Deep Neural Networks}

\author{Yanze Liu, Xuhui Chen, Yanhai Du, Rui Liu$^{*}$ 
    
\thanks{Authors are with The Cognitive Robotics and AI Lab (CRAI), College of Aeronautics and Engineering, Kent State University, Kent, OH 44240, USA. $^{*}$ Dr. Rui Liu is the corresponding author {\tt\small ruiliu.robotics@gmail.com}.}
}

\maketitle
\thispagestyle{empty}
\thispagestyle{fancy}
\setcounter{page}{1}
\pagestyle{fancy} 
\lhead{} 
\chead{} 
\rhead{} 
\lfoot{} 
\cfoot{\thepage}
\renewcommand{\headrulewidth}{0pt}
\renewcommand{\footrulewidth}{0pt}

\vspace{-2cm} 
\setlength{\abovecaptionskip}{-0.1cm}  
\setlength{\belowcaptionskip}{-2cm}

\setlength{\abovedisplayskip}{3pt}
\setlength{\belowdisplayskip}{3pt}

\begin{abstract}

Recently unmanned aerial vehicles (UAV) have been widely deployed in various real-world scenarios such as disaster rescue and package delivery. Many of these working environments are unstructured with uncertain and dynamic obstacles. UAV collision frequently happens. An UAV with high agility is highly desired to adjust its motions to adapt to these environmental dynamics.
However, UAV agility is restricted by its battery power output; particularly, an UAV's power system cannot be aware of its actual power need in motion planning while the need is dynamically changing as the environment and UAV condition vary. 
It is difficult to accurately and timely align the power supply with power needs in motion plannings. 
This mismatching will lead to an insufficient power supply to an UAV and cause delayed motion adjustments, largely increasing the risk of collisions with obstacles and therefore undermine UAV agility. 
To improve UAV agility, 
a novel intelligent power solution, Agility-Enhanced Power Supply (AEPS), was developed to proactively prepare appropriate amount powers at the right timing to support motion planning with enhanced agility. This method builds a bridge between the physical power system and UAV planning. With agility-enhanced motion planning, the safety of UAV in complex working environment will be enhanced. To evaluate AEPS effectiveness, missions of "patrol missions for community security" with unexpected obstacles were adopted; the power supply is realized by hybrid integration of fuel cell, battery, and capacitor. The effectiveness of AEPS in improving UAV agility was validated by the successful and timely power supply, improved task success rate and system safety, and reduced mission duration.

\end{abstract} 


\IEEEpeerreviewmaketitle

\section{Introduction}

With technological advancements in sensor, design, manufacturing, and communication, unmanned aerial vehicles (UAV) is largely developed in perceiving and planning; thus is widely used for applications, such as disaster search and rescue \cite{rescue,rescue1}, border or city patrol \cite{patrol,patrol1}, and commercial UAV shows \cite{stage}. Many of these working environments are unstructured with uncertain and dynamic obstacles. UAV collisions frequently happen, which degrades UAVs' mission performance. An UAV with high agility is highly desired to adjust its motions to adapt to these environmental dynamics.

However, it is challenging to safely deploy an UAV in the real world, as the environmental conditions in terrain and obstacle distributions are dynamically changing. Unexpected situations require immediate motion reactions with timely power supply support. It would be difficult to accurately estimate and align the power supply with power needs in dynamic motion plannings. This mismatching will lead to an insufficient or over-sufficient power supply with undesired robot motion adjustments, further degrading UAV task performance. 
\begin{figure}[t]
	\centering
	\includegraphics[scale=0.5]{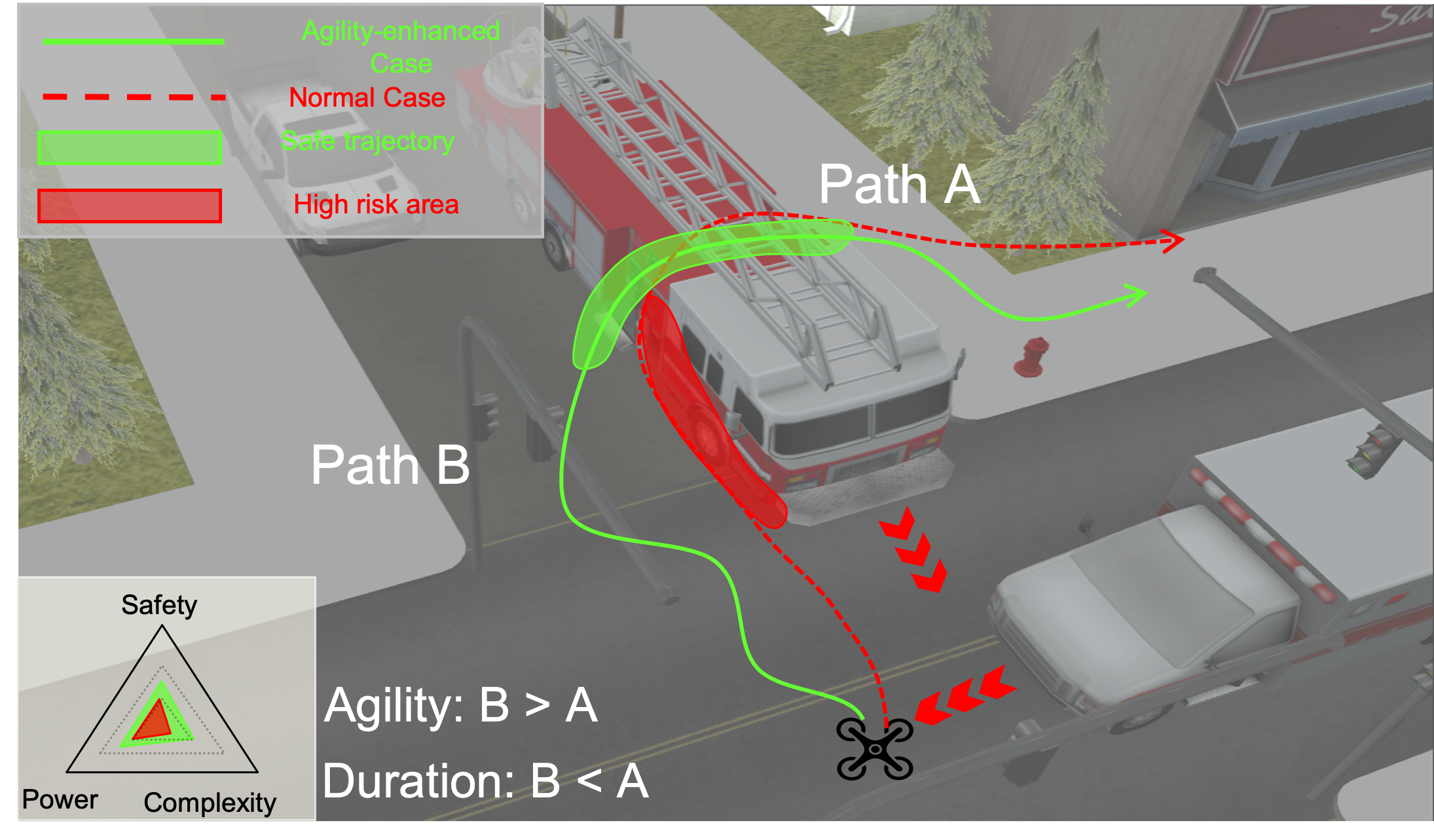}
	\caption{The illustration of improving UAV agility by using AEPS. UAV agility is mainly described from three aspects: \textit{Power}, \textit{Safety} and \textit{Trajectory complexity}. The agility of UAV is improved by using the AEPS.}
	\label{idea}
\end{figure}
First, dynamic obstacles such as building clusters, moving vehicles, and trees inevitably appear in the mission environment. These obstacles constrain the available workspace for UAV deployments; the dynamic appearances of obstacles will further limit the time and space to adjust UAV motions for obstacle avoidance. Second, the UAV system configuration does not support sudden and significant movements such as sharp turns, big accelerations, and emergent obstacle avoidance, which can only be realized on highly coordination between theoretical motion planning and physical systems' motion adjustments. The power supply limitation will cause a mismatch between theoretical planning and action motion realization, further endangering the UAV safety. Third, current UAVs lack the power management capability to consider UAV motion needs and available power supply, thus cannot prepare the physical power system to maximize the motion performance. This limitation further decreases UAV agility, influencing UAV safety in actual missions.

To improve UAV agility, this paper developed a novel intelligent power solution, Agility-Enhanced Power Supply (AEPS). AEPS proactively prepares the appropriate amount of power in advance to support motion planning with enhanced agility. It is used to predict the mission power consumption using the information from the planned trajectory, then pass the prediction to the power supply system for timely charging and discharging of hybrid power sources such as fuel cell, battery, and capacitor.

This model utilizes the power sources' different charging and discharging characteristics, including fuel cell, battery, and capacitor. A fuel cell stores a large amount of electric energy with stable and durable power output, for example, successful application of fuel cell in electric vehicle \cite{fuel,fuel1}; a battery has lower energy storage but with fast and high-peak power output \cite{battery}; an ultracapacitor stores a large amount of power but with instant and high-peak power output \cite{uc}. The power supply contributed by the three types of sources will support but durable UAV flying and emergent situation response. 
With the guidance of the novel power supply model AEPS, the electric energy supply is flexibly allocated to the power sources in advance; so that the battery and capacitor will be charged and the fuel cell will be activated before motion planning. When the UAV performs some actions requiring high-power discharge, the ultracapacitor will supplement the power that the fuel battery cannot provide. Thus, the UAV has enough power output to complete behaviors at a specific moment or provide stable power to maintain stable motion status. The idea is illustrated in Fig.\ref{idea}. This paper mainly has three contributions:
\begin{enumerate}
    \item A novel power-motion relation is developed. The new relation builds the bridge between the UAV power supply and UAV motion adjustments. It lays a solid practical foundation (power supply) for realizing precise motion adjustments by intuitively adjusting the power supply to adjust UAV motions.
    \item A novel intelligent power solution, Agility-Enhanced Power Supply (AEPS), is developed to predict the power consumption of UAV motion plannings. AEPS theoretically enhances UAV agility by timely and accurately estimating power needs of motion adjustments and correspondingly requesting power from the sources. 
    \item  A novel predictive power preparation method is developed based on AEPS. This neural network power prediction method practically enhances UAV agility by coordinating power sources to provide desired power output for UAV motion adjustments.
\end{enumerate}

\section{related work}
UAV agility has been investigated in many recent research. Agility is important for UAVs when avoiding danger from the environment. The meaning of agility can include motion accuracy, output power level, high-speed movement, and reduce the risk of impact et al \cite{l1}\cite{agility3}\cite{agility}\cite{agility4}. \cite{agility3} built a motion planning framework for UAVs to detect surfaces without extra onboard sensors, and the UAV agility got enhanced with the loss of sensor weight. \cite{agility,l1} proposed an adaptive controller for UAV to have more precise control with relatively higher speed to improve agility. In \cite{agility4}, a four-part optimization algorithm was proposed to support real-time flight path tracking and re-computation in dynamic environments for power management. By implementing the method in an intelligent drone management system, it was verified that the insufficient power supply situation happened less. In the above research, however, they only consider the impact of a single element such as motion accuracy, power management, and speed on UAV agility, respectively. Besides, they cannot intelligently adjust the original motion plan according to the highly dynamic environment. These ideas work well in simple situations like building monitory and target tracking in the open field. In our work, the key difference compared with the previous work is that the environment is highly dynamic and sudden, including falling dustbin and fast-moving vehicles. In such an environment, the risk of UAV crash is significantly high because of the common weakness of these methods: they cannot quickly react to dynamic obstacles due to incapability of dynamically calculating the power needs in complex situations and matching the desired power supply. In our research, by predicting the power consumption according to the environment, the AEPS method can dynamically change the original plan to a more agile and safer trajectory with richer safe distances. Besides, compared with improving the agility of UAVs from a single element, improving multi perspectives of agility like power level, UAV speed, and planning trajectory help an UAV to avoid obstacles in a quicker and safer way.

The relationship modeling between path planning and power system in robots had also drawn attention \cite{pp,pp1,k1,pp2}. In \cite{pp,k1}, a time-optimal path planning method considering the total energy of the solar battery was developed. It was realized by detecting the denser solar radiation area in advance and prioritizing the area in solving the planning problem. \cite{pp1} developed a planning technique based on Tabu-search methods to make sure the UAV can avoid running out of power during the mission operation. By comparing the performance with another method, greedy TSP, based planing method, the work showed that the method provided an effective path planning by which a robot can be guaranteed to stay alive and finish all tasks with the minimum energy. Both of the methods consider the influence of insufficient power on UAVs. However, the planning methods did not utilize the power to help UAV with motion adjustments for safety improvements. To address the above problems, a novel power-based robot planning method is developed to adjust the planning route when UAVs are in danger. The planning module and the power system work collaboratively to have a safer trajectory: the planning module plans a route with higher agility to protect UAV from danger. Then, the predicted result is sent to the hybrid power system to support the motion adjustment actively.

\section{Agility-Enhanced Power Supply}
\begin{figure}[t]
	\centering
	\includegraphics[scale=0.47]{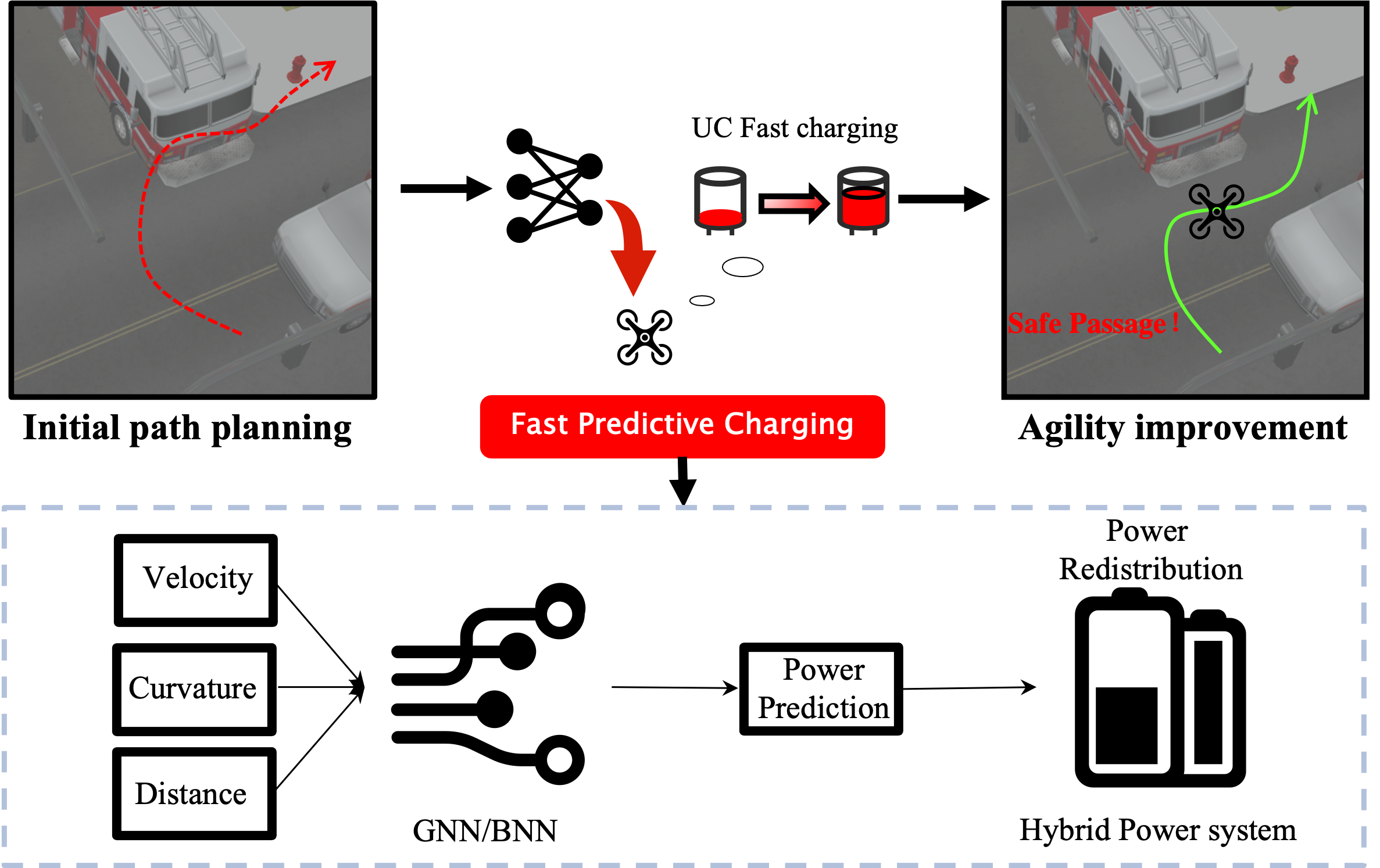}
	\caption{The working mechanism of AEPS to predict the power consumption. First, the information of the planning module is used as the input of the neural network. Then, the ultracapacitor is charged according to the plan. Finally, with the characteristic of high power density, the agility of UAV is improved during the mission.}
	\label{mechanism}
\end{figure}

The Agility-Enhanced Power Supply (AEPS) is developed to provide an intelligent energy distribution strategy to support UAV motion adjustment. As shown in Fig. \ref{mechanism}, the information of the initially planned path is sent to the deep neural network to have a power prediction before the mission operation. Then the power supply system will work collaboratively: the fuel cell will charge the ultracapacitor according to the prediction. During the mission, the UAV will prioritize the capacitor power supply where high power discharge is needed.

\subsection{Power prediction model}
\label{ppm}
There exist power consumption differences in UAV motions. Some motions result in higher power demand. For example, take-off and sharp turns requiring great acceleration require much power. In contrast, hover or smooth driving spends relatively less energy. The difference is the acceleration level: the higher acceleration results in greater power consumption \cite{pa,pa1}. 

A power prediction model is developed based on a deep neural network $M(\theta)$ to estimate the power consumption of motion planning in real-time. $\theta$ denotes the parameter set of the power prediction model. The rationale of modeling the power prediction is that motion adjustments with different action ranges, speeds, durations will consume different power levels. A deep neural network estimates power consumption by analyzing the desired actions, current UAV motion status, and environment influence. The module's input is UAV velocity, planning trajectory length, and mean of the absolute value of the planning trajectory curvature, while the output is the predicted power consumption. The parameter set is composed of multiple quantities. It incorporates the velocity, trajectory complexity, and length of the trajectory in \cite{formular}. Concretely, the curvature and length represent the trajectory complexity and the mission duration. These parameters are the inputs of the $M(\theta)$, while the output is the power prediction of the mission. The training label is $\hat{P}$ according to the model. The $M(\theta)$ is used to explore the nonlinear relation between the physical quantities of UAV operation and the power consumption in the actual case.
Concretely, actions such as sharp turns and sharp rises will significantly improve the power prediction level. The original planning information-power relation modeling process is illustrated in \cite{formular}, and the previous power prediction is $\Tilde{P}$. More elements are incorporated from the planning module to form the algorithm mechanism. The novel prediction model is
\begin{align}\label{model}
    P = \Tilde{P} + 10^{-4}|C| + 2\times 10^{-5}D\Bar{|C|}
\end{align}
where $\bar{|C|}$ is the mean of the absolute value of trajectory curvature, while $|C|$ is the absolute value in the corresponding second. $D$ is the length of the trajectory. This model incorporates the information from the planning trajectory. When the UAV needs to avoid horizontally moving obstacles, such as moving cars, the agility-enhanced UAV will accelerate during the obstacle avoidance process. Meanwhile, the trajectory will change to allow the UAV to stay far away from the obstacle. The curvature of the changed route is greater than the original one. The power demand is promoted due to the improved curvature and velocity. The most dangerous scenario is to avoid falling objects, which are usually characterized by fast and irregular movement. Such characteristics require the UAV adjust its action quickly. Once a falling object is detected, the agility-enhanced UAV will accelerate away in the opposite direction of the object. The key to avoiding a crash is the great UAV acceleration, resulting in more power consumption.

\subsection{Smart Power Supply for UAV Motion Agility Enhancement}

As shown in Fig.\ref{mechanism}, a learning framework that utilizes a neural network to predict the power consumption of the mission is developed. The prediction guides the hybrid power system to distribute the power in advance. First, the power prediction model $M(\theta)$ predicts the power consumption before the mission operation. Then the output is sent to the hybrid power system to assign the power distribution. The prediction values produced by the model (\ref{model}) are considered as the training label. The goal of the learning model is to solve the optimization problem as below:
\begin{align}
    \min_\theta L(M(\theta))
\end{align}
where the performance of the learning model is described by calculating the L2-Norm loss of $P$ and $\hat{P}$:
\begin{align}
    L(M(\theta))=\frac{1}{2}\Vert P-\hat{P}\Vert
\end{align}
which is also the definition of the cost function. Predicting the power consumption is viewed as the regression problem. It is used to approximate the real power demand. For each training epoch, a batch of $n=1100$ planning results is used. The results include all the elements in the model (\ref{model}). Each result gives a new power label $p_i(i=1,...,n)$. The parameters of $M(\theta)$ are updated as:
\begin{align}
    \theta'=\theta-\alpha\nabla_\theta L_{D_{train}}(M(\theta))
\end{align}
where $\alpha$ is the learning rate of the optimization problem, and $D_{train}$ is the training data set. The validation set $D_v$ is used to validate the training result.

\begin{figure}[t]
	\centering
	\includegraphics[scale=0.39]{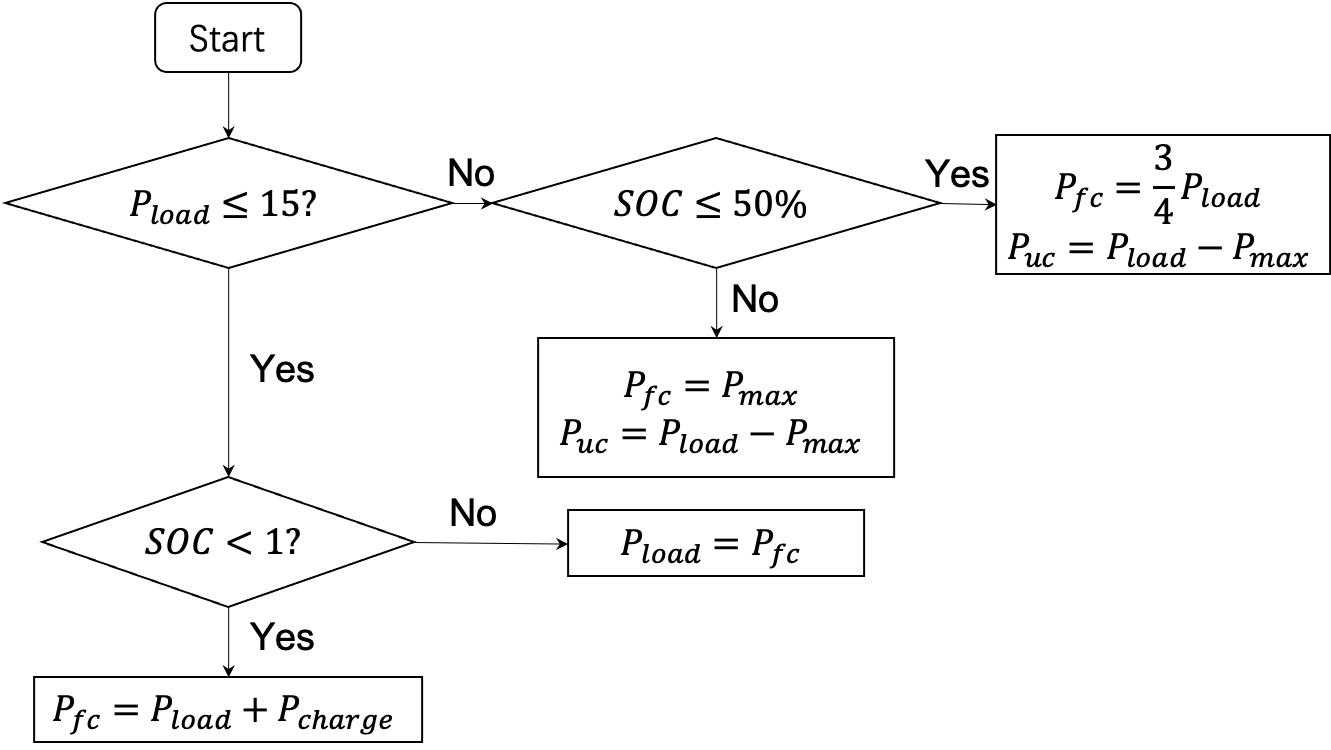}
	\caption{The flow chart of the rule-based method to allocate the power distribution. The strategy guides the hybrid power system how to function.}
	\label{algorithm}
\end{figure}
\subsection{Power supply preparation}

The power supply system comprises battery, fuel cell, and ultracapacitor. The fuel cell and battery characteristic is high capacity but low discharge power. On the contrary, relatively low capacity but high discharge power are the characteristics of the ultracapacitor. Electric energy could be redistributed according to the prediction $P$ using the complementary characteristics of the three components. Here, a Rule-Based framework is designed as Fig.\ref{algorithm}. It allocates the power distribution to different power sources. In this flow chart, $P_{load}$, $P_{fc}$, $P_{uc}$ represent the power required for the mission, output power of fuel cell and battery, and output power of ultracapacitor, respectively. $P_{max}$, $P_{charge}$, $SOC$ are the maximum output power of fuel cell and battery, charging power of fuel cell and battery to ultracapacitor, and the state of charge of the ultracapacitor. The idea of the hybrid power supply system is to prioritize the fuel cell and battery to supply. The ultracapacitor shall make up the missing electric power. Meanwhile, once the action power requirement is lower than the cell max discharge power, the cell will charge the ultracapacitor if it is not fully charged. In the hybrid power system, the ultracapacitor's power output and remaining power influence some critical moments like avoiding unexpected obstacles. The more power remaining and the greater the output power, the more aggressive the route assigned by the planning module, so as to be more likely to stay away from danger.

\section{Experiment and evaluation}

In this work, two UAV patrol missions in community emergency were designed to validate the effectiveness of AEPS. The UAV missions were to find the cause of the community emergency, such as fire or traffic accident. The mission requires to complete the patrol and ensure the safety of the UAV. Here are three aspects that were evaluated. (\romannumeral1) the effectiveness of AEPS in improving UAV agility, which is composed of the UAV safety, trajectory complexity, and power level. (\romannumeral2) the effectiveness of AEPS in reducing the mission duration. (\romannumeral3) the prediction accuracy comparison between Bayesian neural network (BNN)\cite{bnn} and Graph Neural Network (GNN)\cite{gnn}.

\begin{figure}[t]
	\centering
	\includegraphics[scale=0.54]{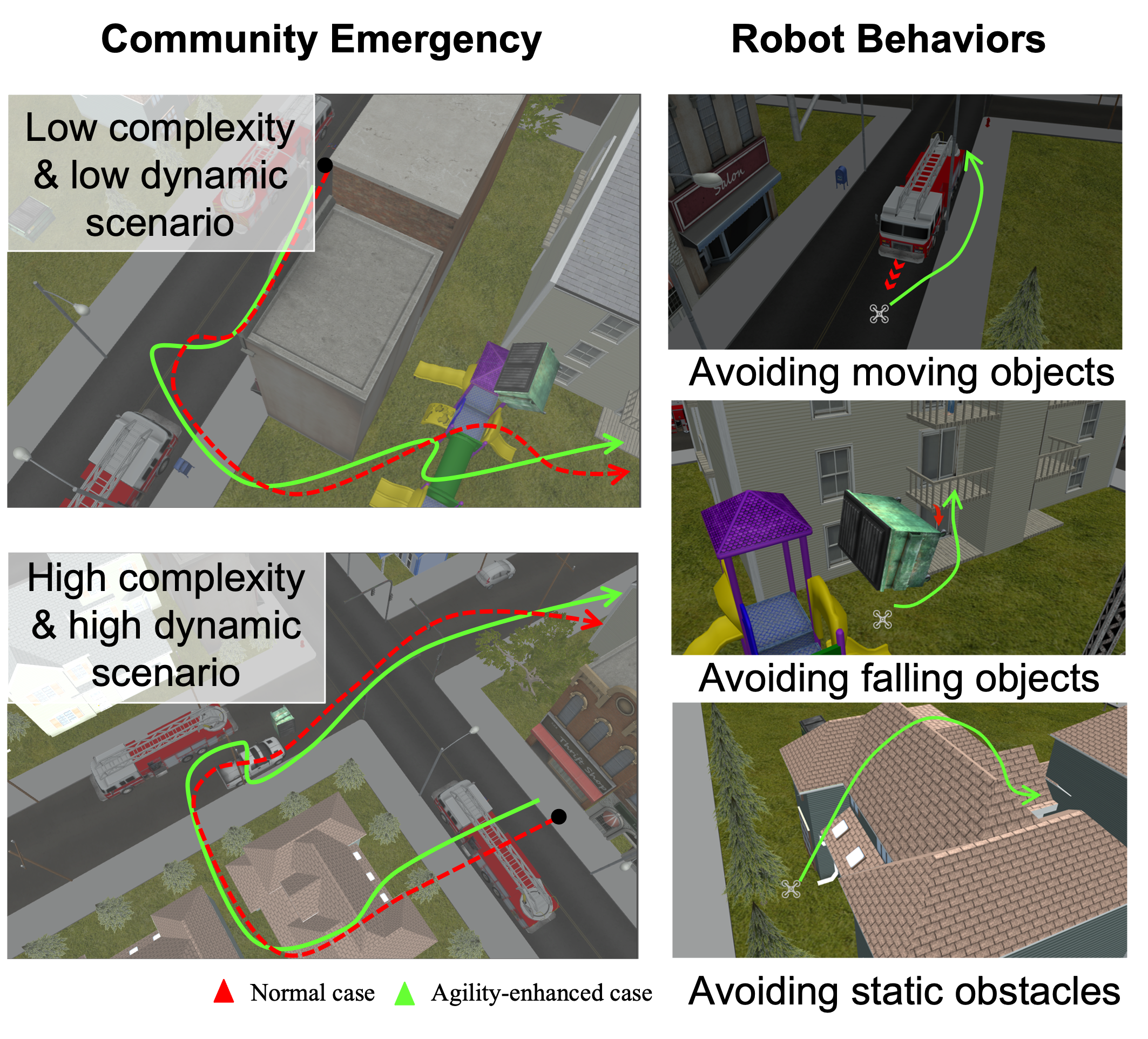}
	\caption{Illustration of the experiment setting, include two UAV patrol mission in community emergency, the agility comparison of normal case and agility-enhanced case and the UAV behaviors. Two cases, agility-enhanced case and normal case, are investigated.}
	\label{mission}
\end{figure}
\subsection{Experiment setting}

Fig.\ref{mission} shows the experiment environment "UAV patrol mission in community emergency" by using the lab simulation platform CRAImrs. CRAImrs is a simulation platform to simulate human supervisory robots and UAVs deployments in real-world scenarios by using latest artificial intelligent algorithms; CRAImrs is developed and introduced in our previous works \cite{h,y,ppp}.
The mission location is set to $300m\times 300m$ in a community. There are two scenarios: \textit{low complexity and low dynamic scenario} and \textit{high complexity and high dynamic scenario}. The UAV was required to avoid moving vehicles, falling dustbin to reach the goal in the first scenario. While, in the second scenario, the UAV was required to reach the destination after avoiding other UAVs and avoiding moving vehicles and falling objects at the same time. The single UAV searched the emergency cause in the community like abnormal fire or criminal. In this environment, there is a dynamic risk such as the static obstacles and moving vehicles and unpredictable factors like the falling items. 

\begin{figure}[t]
	\centering
	\includegraphics[scale=0.3]{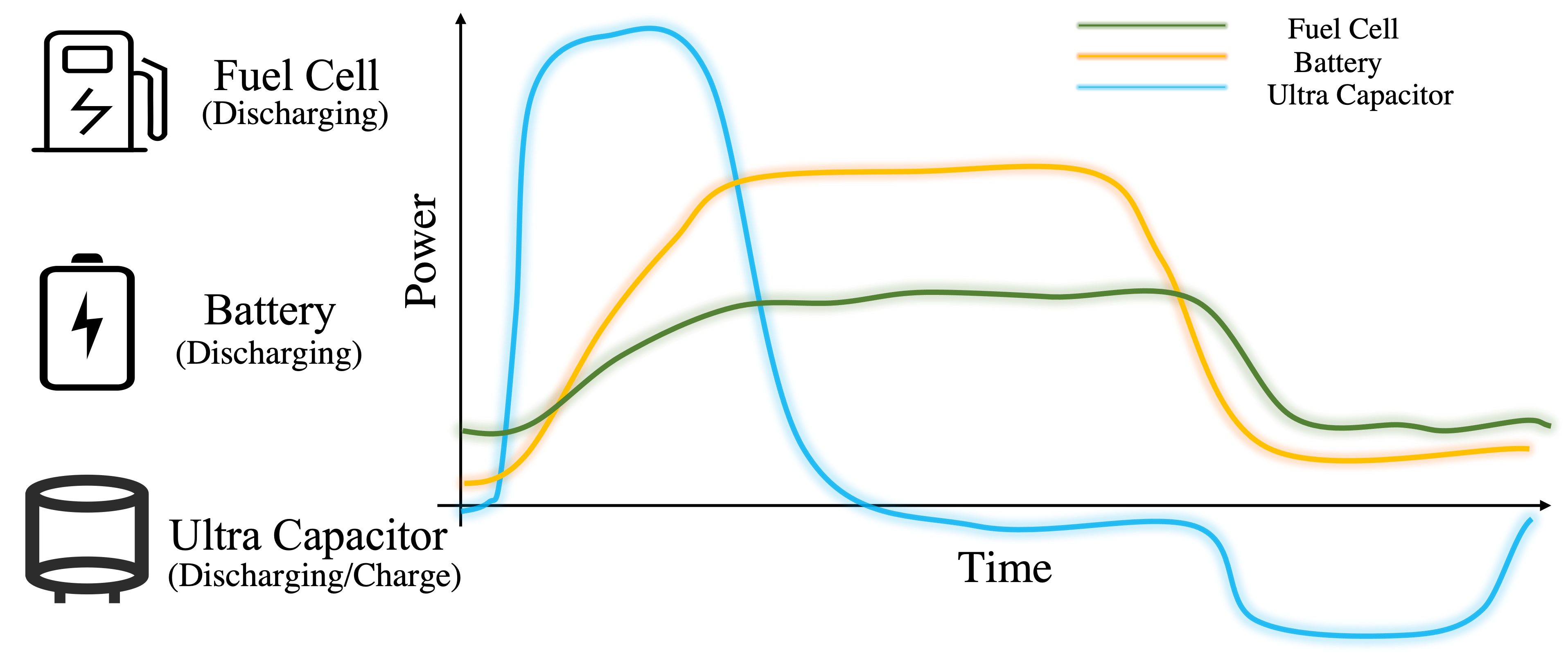}
	\caption{Illustration of the power distribution of different power source. The UAV power demand in the specific moment can be fulfilled by utilizing the characteristic of different power source. }
	\label{source}
\end{figure}

\begin{figure}[t]
	\centering
	\includegraphics[scale=0.36]{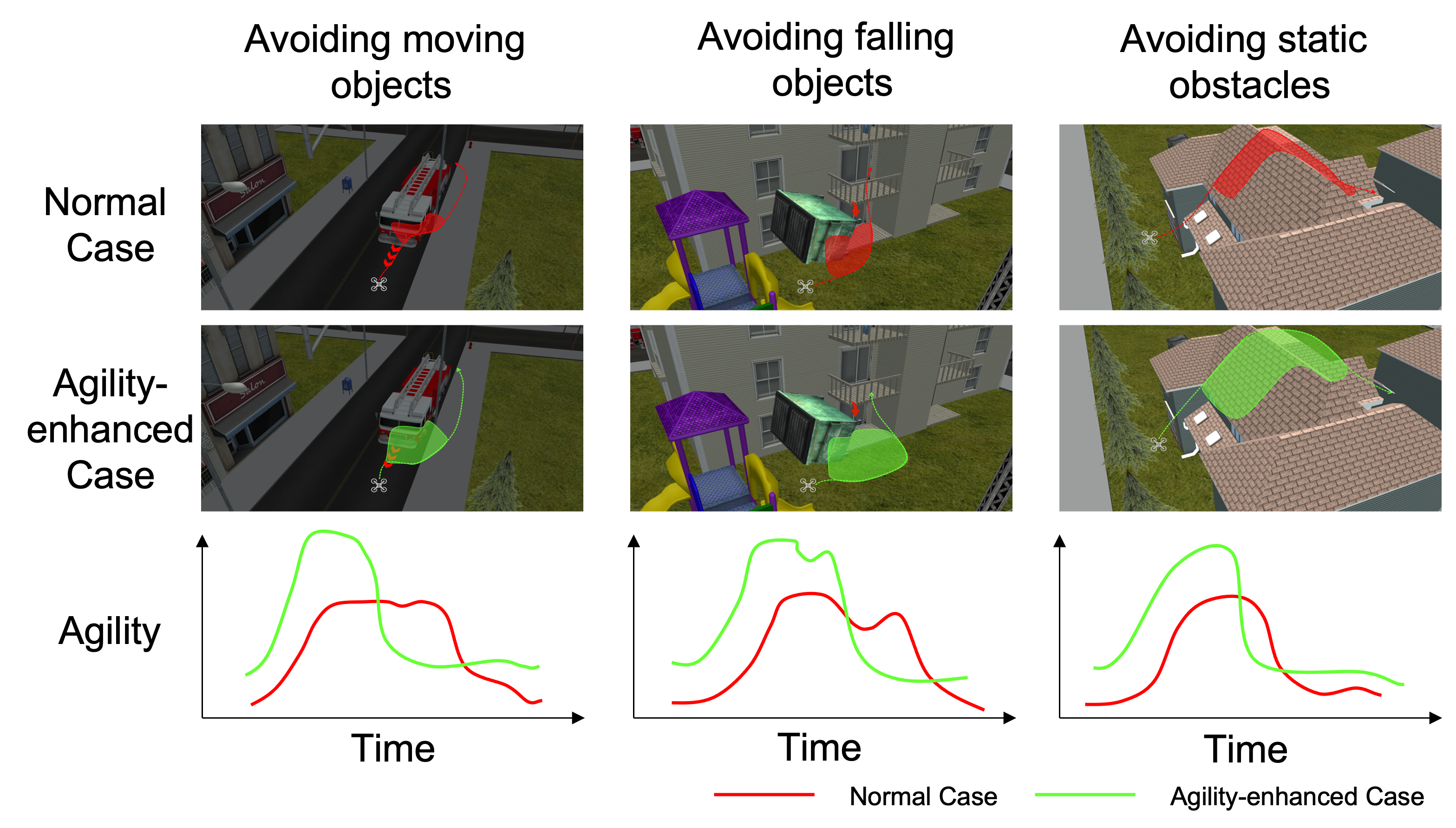}
	\caption{The comparison of UAV behavior difference and agility for agility-enhanced case and normal case}
	\label{illu}
\end{figure}

Thus, the action modes of UAV in obstacle avoidance were: 1). "Distance between UAV and obstacles" represented by minimal gap between UAV and obstacles was set with in [1m, 3m]; 2). "UAV trajectory curvature" denoted by the average curvature of the UAV trajectory was set with in (0$m^{-1}$, 1$m^{-1}$ ]; 3). "UAV flying speed" defined by the average speed of UAV was set within [3m/s, 8m/s]; 4)."UAV flying distance" represented the total distance the UAV traveled was set within [30m, 45m]. For the patrol tasks, the agility of the UAV was described in three aspects: "\textit{power}", "\textit{safety}" and "\textit{trajectory complexity}". "\textit{power}" means the average power level of the UAV during the mission. "\textit{safety}" represents the distance between the UAV and the objects in the environment. "\textit{trajectory complexity}" represented the variance of the trajectory curvature. Also, to validate the effectiveness of AEPS, two cases, normal case and agility-enhanced case, were developed. The UAV in normal case used the traditional method, while, in agility-enhanced case, the UAV utilized the AEPS method to complete the mission. The UAU motion performance under different cases are illustrated in Fig.\ref{illu}. Besides, the power level of different power source is illustrated in Fig.\ref{source}, which shows the power changes of different power sources when performing
tasks. The specific configuration of the power system is the Li-ion-based ultracapacitor is 1000F, and the maximum power output is 30W. The battery's maximum power output is 13W, and its capacity is 2000mAh. The fuel cell's capacity is 10000mAh, while the maximum power output is 3W. The UAV weighs 900g and has a size of 347 mm × 283 mm × 107 mm. Besides, it is equipped with a 3D lidar to detect obstacles. The maximum payload of the UAV is 600g. To analyze the performance of AEPS when using different neural networks, two kinds of the neural network are developed, BNN and GNN, to see which one can better predict the actual situation. The duration of each patrol mission is compared and analyzed.

\subsection{Result analysis}

\begin{figure}[t]
	\centering
	\includegraphics[scale=0.34]{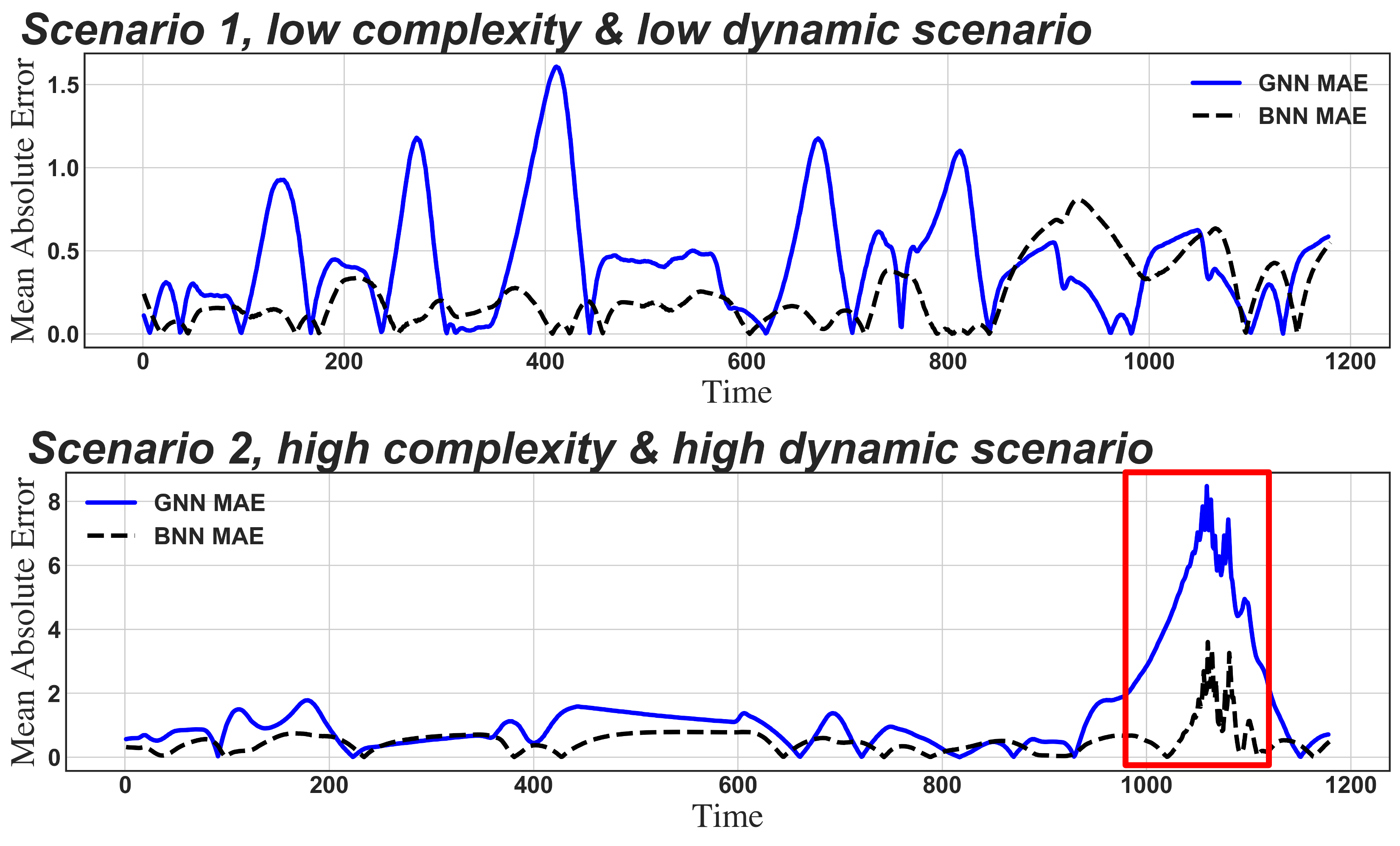}
	\caption{The mean absolute error of two neural networks, BNN and GNN, between the prediction and the power consumption in real case. The comparison shows that the prediction of BNN has the smaller error than GNN's.}
	\label{mae}
\end{figure}

\textbf{Comparison of two neural networks.} In order to achieve the best performance of AEPS, two kinds of neural networks, BNN and GNN, were developed to compare the accuracy of their predictions. Based on the learning framework in \ref{ppm}, the mean absolute value (MAE) of the two neural networks was measured for two patrol missions. MAE showed the absolute difference value between the prediction and the actual power consumption. The result was shown in Fig.\ref{mae}, in which the mean absolute value of BNN was smaller than GNN's. The average maximum error of BNN was $58.34\%$ of GNN. Besides, in terms of robustness, at the most intense moment illustrated in the red box of the Fig.\ref{mae}, the error of BNN was $31.25\%$ of GNN's. In conclusion, BNN performs more robustly than GNN in the current setting. The probabilistic nature of BNN is the source of the robustness of BNN based AEPS.

\begin{figure}[t]
	\centering
	\includegraphics[scale=0.36]{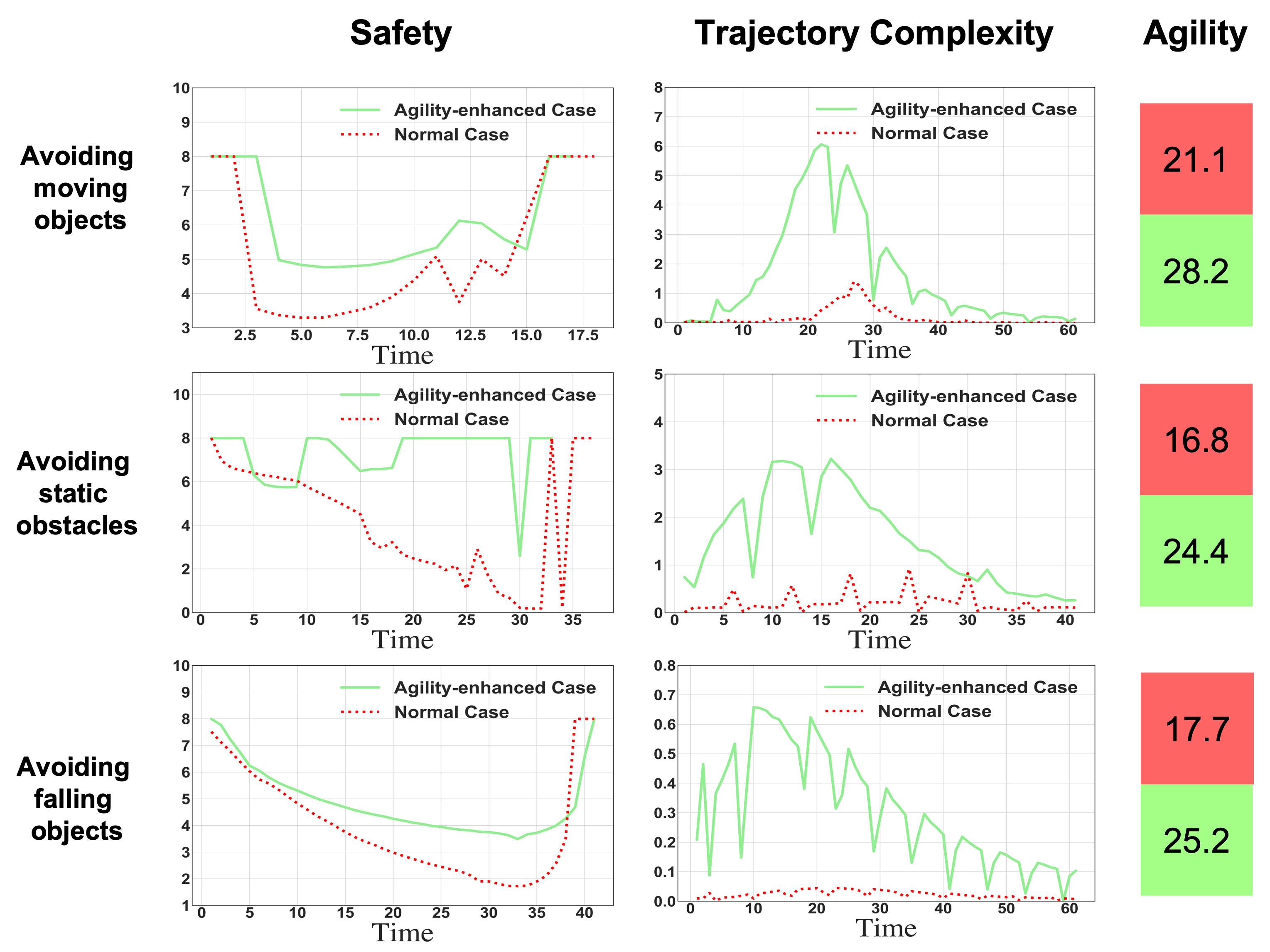}
	\caption{Illustration of the performance and agility comparison for specific robot behaviors between normal case and agility-enhanced case. "\textit{Safety}" is the distance between UAV and obstacles. "\textit{Trajectory Complexity}" is the curvature of the trajectory.}
	\label{spe}
\end{figure}

\begin{figure}[t]
	\centering
	\includegraphics[scale=0.37]{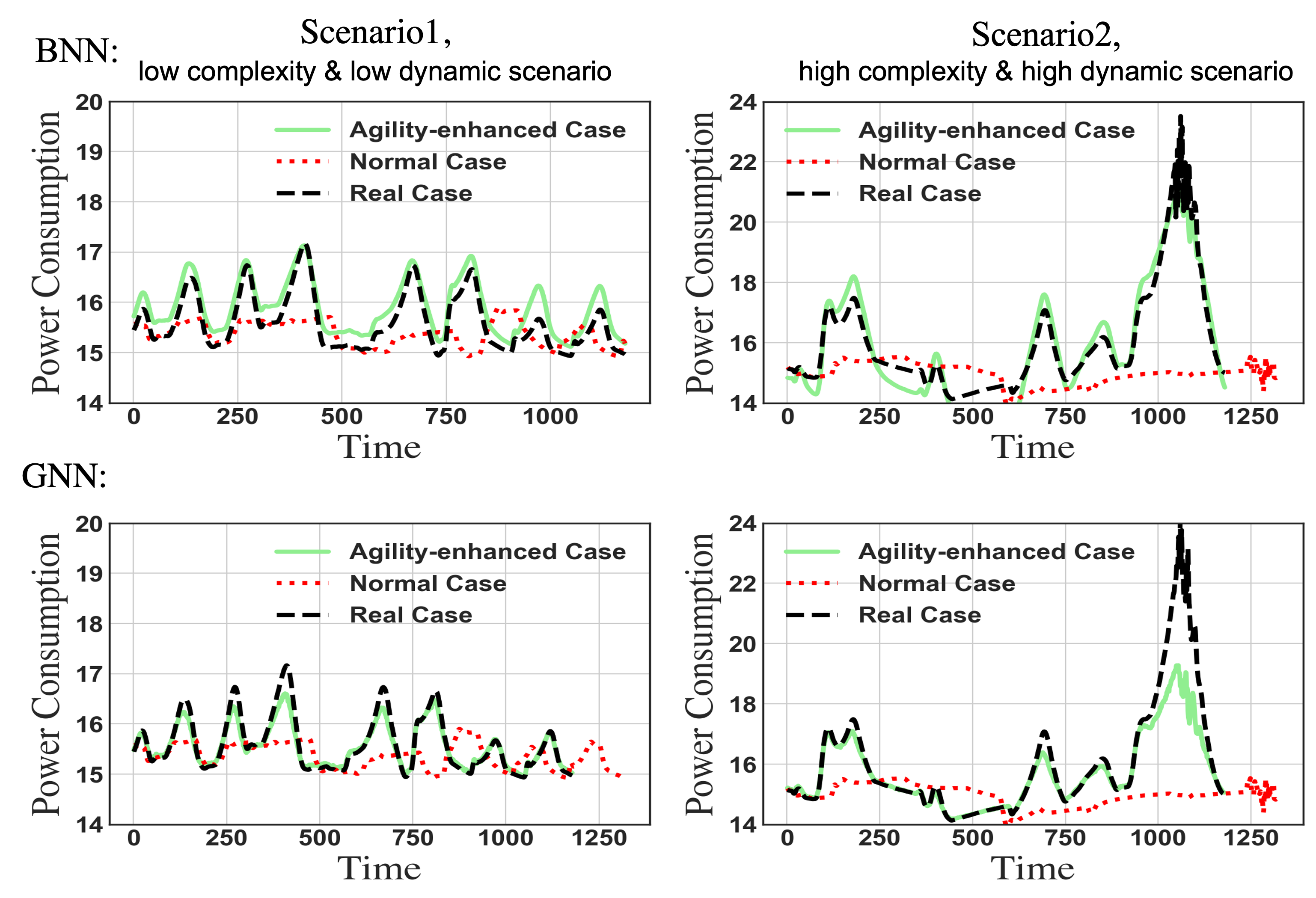}
	\caption{For the illustration of BNN in the first row, it shows the power comparison between the power consumption in normal case, power consumption in agility-enhanced case. For GNN in the second row, it shows the power comparison between the power consumption in normal case, power consumption in agility-enhanced case and its corresponding prediction.}
	\label{power}
\end{figure}

\textbf{Agility Improvement Evaluation.} The agility of UAV was described by: the UAV \textit{power} $P$ and the UAV performance containing \textit{safety} $S$ measured by the distance between UAV and obstacles, and \textit{trajectory complexity} $C$ measured by the UAV trajectory curvature. Agility $A$ was defined as:
\begin{equation}
    A=P+C+S
\end{equation}
where safety was the minimum, while other quantity took the maximum of the whole process in above equation. Here, we used the maximum agility to represent the whole process. The overall performance and the specific movement of the UAV were analyzed. For the specific robot behaviors, there were three types motions: avoiding moving objects, avoiding the static obstacles and avoiding the falling objects. The changes in safety and trajectory complexity for these actions were shown in the Fig.\ref{spe}. The agility was also valued for the two cases. The results were: 1) Compared with the normal case, in the agility-enhanced case, the safety level has an average increase of $58.16\%$. 2) Compared with the normal case, the trajectory complexity has an average increase of $84.86\%$ in the agility-enhanced case. 3) Compared with the normal case, the UAV agility has an average increase of $40.25\%$ in the agility-enhanced case. For the overall performance, the power prediction using two neural networks, i.e. BNN and GNN, was plotted to validate the AEPS in Fig. \ref{power}. The power prediction of the learning module, the power consumption in normal case, and the real power consumption were illustrated. The average real-time power consumption was shown in Fig.\ref{power}. The trajectory complexity in agility-enhanced case scene was $94.57\%$ higher than that in normal scene in the first scenario, while in the second scenario the number was $44.13\%$. The average curvature of agility-enhanced case ($1.53$ and $7.60$) was $47.19\%$ greater than that of the normal case ($0.08$ and $4.25$). The safety level in agility-enhanced case scene ($3.18$) was $9.84\%$ higher than that in normal scene ($2.95$) in the first scenario, while in the second scenario the number was $93.21\%$ ($2.56$ versus $0.20$). It shown that the AEPS can greatly improve the UAV agility because it concerns multiple aspect including power, obstacle distance, and trajectory curvature that decide the UAV agility.

\textbf{Response Time Improvement.} In this work, the mission duration was analyzed to validate the effectiveness of the AEPS in reducing mission time. The average duration of the ten experiments was calculated in each scenario. The completion time in agility-enhanced case scene was $10.45\%$ shorter than that in the normal scene in the first scenario, while in the second scenario, the number was $9.05\%$. From the distribution diagram, in the first scenario, the median duration was $1.97$ for the agility-enhanced case, while it was $2.18$ for the normal case. In the second scenario, the median duration was $1.99$ for the agility-enhanced case, while $2.19$ for the normal case. There was a $10\%$ reduction in running time on average. It proved that UAV that utilizes the AEPS method can reduce the mission duration according to the analysis data.

\section{Conclusion}

This paper developed a novel UAV agility improvement method, Agility-Enhanced Power Supply (AEPS), to enable the unmanned aerial vehicle to gain higher agility by distributing the power intelligently in a hybrid power system. Two UAV patrol missions in community emergencies were developed to validate the method's effectiveness. Each mission was run on two cases, \{agility-enhanced and normal cases\}. The effectiveness of the novel method for improving the UAV agility was validated by the improved agility and the reduced mission duration. Given the capability of improving the power output to improve motion planning accuracy, AEPS is beneficial for general robotic systems for safe trajectory planning. In the future, research on upgrading the hybrid power system could be done to gain a faster response, such as the increase of the charge of the ultracapacitor, for better agility enhancements.


\ifCLASSOPTIONcaptionsoff
  \newpage
\fi

\bibliographystyle{ieeetr}

\bibliography{bibtex.bib}

\end{document}